\documentclass[10pt,journal,compsoc]{IEEEtran}

\ifCLASSOPTIONcompsoc
  \usepackage[nocompress]{cite}
\else
  \usepackage{cite}
\fi

\ifCLASSINFOpdf
\else
\fi

\usepackage[pdftex]{graphicx}
\usepackage{enumitem}
\setlist[enumerate]{itemsep=0mm}

\usepackage{algorithm2e}
\usepackage{color}
\usepackage{subfigure}
\graphicspath{{figures/}{pictures/}{images/}{./}} 

\usepackage{cite}
\usepackage{graphicx}
\usepackage{amsmath}

\usepackage{booktabs} 
\usepackage{multirow}
\usepackage{array}
\usepackage{caption}

\usepackage[table]{xcolor}

\begin{document}

\title{Forecasting People's Needs in Hurricane Events
\\from Social Network}

\author{Long~Nguyen,
        Zhou~Yang,
        Jia~Li,
        Zhenhe Pan,
        Guofeng~Cao,
        and~Fang~Jin
\IEEEcompsocitemizethanks{
\IEEEcompsocthanksitem Long Nguyen, Zhou Yang, and Fang Jin are with the Department of Computer Science, Texas Tech University.
E-mail: \text{\{long.nguyen, zhou.yang, fang.jin\}@ttu.edu}
\IEEEcompsocthanksitem Jia Li is with the Department of Civil, Environmental, and Construction Engineering, Texas Tech University. Email: jia.li@ttu.edu. 
\IEEEcompsocthanksitem Zhenhe Pan is a senior software engineer at Kinetica db Inc. Email: zhenhepan@gmail.com.
\IEEEcompsocthanksitem Guofeng Cao is with the Department of Geosciences, Texas Tech University. Email: guofeng.cao@ttu.edu.
}

}

\IEEEtitleabstractindextext{%
\begin{abstract}
Social networks can serve as a valuable communication channel for asking for help, offering assistance, and coordinating rescue activities in disaster because it allows users to continuously update critical information in the fast-changing disaster environment. This paper presents a novel sequence to sequence based framework for forecasting people's needs during disasters using social media and weather data. It consists of two Long Short-Term Memory (LSTM) models, one of which encodes input sequences of weather information and the other plays as a conditional decoder that decodes the encoded vector and forecasts the survivors' needs. Case studies using data collected during Hurricane Sandy in 2012, Hurricane Harvey and Hurricane Irma in 2017 demonstrate that the proposed approach outperformed the statistical language model n-gram, LSTM generative model, and convolutional neural network (CNN) based model. This research indicates its great promise for enhancing disaster management such as evacuation planning and commodity delivery.

\end{abstract}

}

\maketitle

\IEEEdisplaynontitleabstractindextext

\IEEEpeerreviewmaketitle

\ifCLASSOPTIONcompsoc
\IEEEraisesectionheading{\section{Introduction}\label{sec:introduction}}
\else
\section{Introduction}
\label{sec:introduction}
\fi

\IEEEPARstart{S}{ocial} media allows messages to propagate quickly through a large population, facilitating the communication of people's needs and resource availability during disasters and thus ultimately enhancing the effectiveness of disaster relief efforts~\cite{gao2011harnessing}. With social media, everyone can share what they see or hear to create a comprehensive multi-faceted view of critical events that are continually updating as conditions change. This is especially true when disaster happens. During and after disasters, people tend to engage more actively with social media to learn and/or report the latest information. 
Fraustino et. al.~\cite{fraustino2012social} presented a number of case studies and discussed different scenarios when social media can be used to support disaster relief efforts. For example, after the 2011 Japanese tsunami, more than 5,500 tweets were posted every second concerning the disaster~\cite{fraustino2012social}; in the 2011 Haiti earthquake, Twitter was reported as one of the primary means for people to communicate with each other.
The unique role of social media in disasters has motivated stakeholders such as Department of Transportation, Blue Cross Organization, and various telecommunication companies to monitor social media posts to help them take timely actions to mitigate the impacts of disasters. 

From various perspectives (engineering, social, and political), it is imperative to identify ways to address the adverse consequences of disasters and promote long-term recovery~\cite{masten2008disaster}. Many groups have conducted research on facilitating disaster communication and operations using social media~\cite{austin2012audiences, houston2015social, williams2012use}. Recent stories and reports describing the successful application of social media in disaster relief have sparked congressional interest, leading to discussions of how social media can be used to improve disaster response and recovery capabilities at both federal and state levels~\cite{dufty2012using}.

\begin{figure}[t]
    \centering
    \subfigure{
        \includegraphics[scale=0.4]{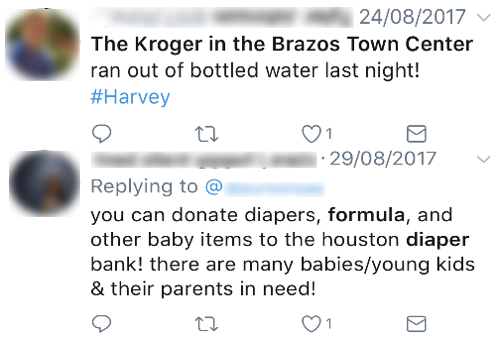}
        \label{sampletweet-tweet}
    } 
    \subfigure{
        \includegraphics[scale=0.2]{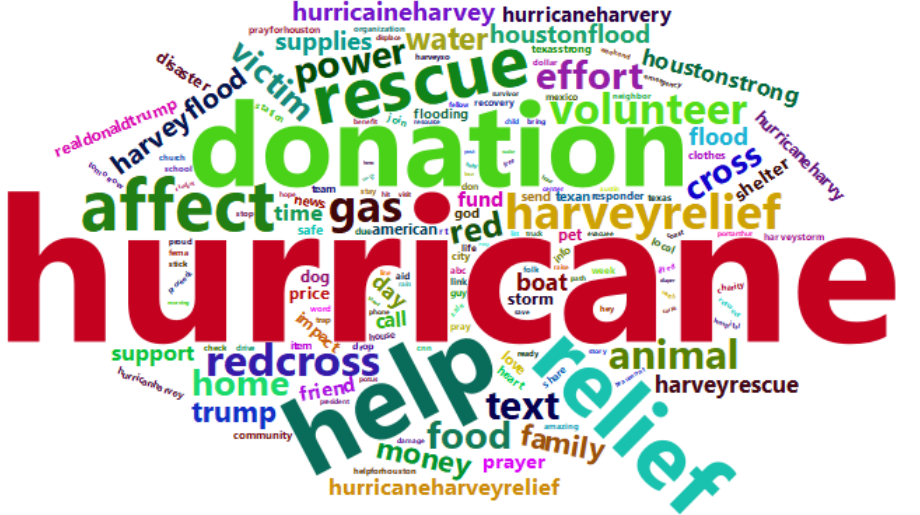}
        \label{sampletweet-wordcloud}
    } 
    \vspace{-3mm}
    
    \caption{Sample tweets to report needs issues and a word cloud showing Tweets discussion during Hurricane Harvey.}
    \label{fig:needs_example}
\end{figure}

This paper aims at developing a framework to forecast people's needs using both weather data and social network data. To survive disasters, people make various preparations beforehand, including storing extra food, water, and gasoline, and boarding up windows and securing water heaters and electric devices. These actions are intended to reduce the risk of injury and reduce the potential damage within individual households caused by potential interruption of lifeline infrastructures (power grids, water supply networks, transportation networks, etc.)~\cite{paton2003disaster}. At system level, understanding people's needs enhances planning and rescue operations and improves community resilience. Before Harvey landed in Houston, for example, tweets about the estimated impacts were the hottest topics on Twitter with many hashtags, there were more than six thousand tweets with the hashtag `\#harvey' on August 25, 2017, with people tweeting about stocking up on food and other necessities to survive the upcoming hurricane. While more information about Hurricane Harvey was being disseminated on social media, people flocked to supermarkets to purchase water, milk, diapers, batteries, flashlights, and first aid items, and as a consequence, most of these necessities in Houston were quickly out of stock. If people's needs can be forecast more accurately with a reasonable horizon (lead time), commodity flows can then be arranged in advance, thus mitigating such shortages. Due to rapid urbanization, the number of people living in disaster-prone areas has grown quickly; it has become increasingly important to be able to conduct reliable and timely assessments of emergency preparedness for impending disasters~\cite{Mandel2012}. The accurate prediction of disaster-related needs would not only promote public preparedness, but also enhance the preparedness of entities such as the government, NGOs, and profit-making organizations, enabling them to work together to tackle disasters collaboratively.

The prediction of disaster-related needs has attracted increasing attention from researchers. A study conducted by Bayleyegn et al.~\cite{BayleyegnTesfaye2006RAot} used a survey to collect information about house damage, illness/injury, and access to utilities. They concluded that the top needs are mental health and primary care services, information about safe generator use, and ways to access medical care and medications. Tahora et al.~\cite{NazerTahoraH.2017IDRv} summarized that the success of a disaster relief and response is mainly dependent on timely and accurate information regarding the status of the disaster, the surrounding environment, and the affected people. Though many researchers have explored the issue of disaster relief, there are limited studies focusing on forecasting of people's needs in a disaster from social networks.

This paper will exploit the rich disaster need-related information contained in social media data. We collected data during Hurricane Sandy in 2012, Hurricane Harvey in 2017, and Hurricane Irma in 2017. A few simple real-world examples can illustrate the vast potential of this approach. Figure \ref{fig:needs_example} shows that during the Hurricane Harvey, a user tweeted that Kroger in the Brazos Town Center ran out of water, along with a word cloud representing the most frequently mentioned terms. When thousands of tweets like this one are collected, analyzed, and projected, a better insight into the collective needs of people will be gained. Among other potential applications, accurate prediction of disaster-related needs is of great importance. 

We will use Hurricane Harvey in our case study. After Hurricane Harvey made landfall, it dumped trillions of gallons of rain on regions of Texas and Louisiana and caused unprecedented flooding. With the majority of the city flooded, and most of the stores closed, people's needs for basic necessities could no longer be met through their typical methods. People in need of basic goods could not buy them from stores or online. By posting their demands for goods on social media (see Figure \ref{fig:needs_example}), many people received help from friends or strangers nearby. Using Hurricane Harvey as a case study will allow us to gain a timely understanding of how social network data can serve the purpose of disaster relief, in particular, for forecasting people's needs ahead of time.

This paper proposes a novel systematic framework to forecast people's needs in disasters using social media data. The dataset consists of tweets from three destructive hurricane events collected from Twitter: Hurricane Sandy, Hurricane Harvey and Hurricane Irma. By filtering keywords, we scraped approximately $150,000$ tweets between October 24, 2012 and November 02, 2012 for Hurricane Sandy, $140,000$ tweets between August 23, 2017 and September 02, 2017 for Hurricane Harvey, and $60,000$ tweets between September 04, 2017 and September 13, 2017 for Hurricane Irma.
A prediction model was built by integrating techniques of recurrent neural networks. Notably, we introduce the sequence to sequence approach in the domain of machine translation for need forecast, which is the first attempt in this direction. We have experimented with the Hurricane events dataset using this new analysis and forecast framework. The main contributions of this paper can be summarized as follows:

\begin{itemize}[noitemsep,nolistsep]
    \item To the best of our knowledge, this study is the first one dedicated to forecasting of disaster-related needs using massive online social media data and weather data. Fundamentally, this unusual discovery opens a gateway for employing online social media data for disaster commodity management.

    \item We introduce the sequence to sequence forecasting model from machine translation domain. The forecaster is trained to maximize the conditional probability of a target sequence given a source sequence. We also design an attention mechanism in the forecasting model to better identify and capture important variables in the input source sequence.
    
    \item By integrating spatial-temporal features, our forecasting model is able to uncover peoples' need trends and geographical distribution in real time, which has the potential to improve the efficiency of disaster management substantially.
    
\end{itemize}

\section{Related Work}
\label{sec:related-work}
\subsection{Disaster Relief with Social Media.}
There is a large body of work using social media data for disaster relief. 
Gao et al. built a crowdsourcing platform to provide emergency services in the 2010 Haiti earthquake, such as food requests\cite{gao2011harnessing}. They integrated the system with crisis maps to help organizations identify the location where supplies are most needed.
The Federal Emergency Management Agency~\cite{lindsay2011social} has used social media to facilitate polls, query the affected, and manage incoming messages from the user-base to help with recovery efforts. Their service includes communicating with the users about emergencies and warnings, providing assistance for requests, and keeping the public updated about the current state of the disaster. \cite{du2019twitter} compared people's concern flow between Twitter and news during California mountain fire.
\cite{lu2015visualizing} explored the underlying trends in positive and negative sentiment with respect to disasters and geographically related sentiment using Twitter data.
Assessment of disaster damage was also investigated at \cite{kryvasheyeu2016rapid}.

\cite{imran2013practical} worked on the extraction of reliable disaster-relevant information from social media to enhance scientific inquiry and accelerate the building of disaster-resilient cities. \cite{yang2017harvey} presented rescue scheduling algorithms during Hurricane Harvey.
\cite{varga2013aid} also proposed a system to facilitate communication between victims and humanitarian organizations in large scale disaster situation. A recent survey from Nazer et. al. ~\cite{nazer2017intelligent} summarized that, the success of a disaster relief and response process relies on timely and accurate information regarding the status of the disaster, the surrounding environment, and the affected people. To be specific, Castillo et. al.\cite{castillo2016big} pointed that \textit{"social media can contribute to situational awareness during crisis, but handling its volume and complexity makes impractical to be directly used by analysts."}. Reuter et. al. brought together the use of social media in emergencies after fifteen years of development with special emphasis on usage patterns, role patterns and perception patterns\cite{reuter2018fifteen}.
Most of the earlier work is mainly focusing on assisting current situation via dissemination of information, optimizing collaboration of volunteers and other authorities, and providing support based on actual requested needs in critical time manner. As a comparison, our paper tackles disaster relief in a different angle by forecasting people's concerns and needs which can assist volunteers, citizens and authorities to prepare disaster relief work ahead of time.
\vspace{-3mm}
\subsection{Sequence to Sequence Model}
Many learning activities require the output to possess a sequential pattern. Similarly, other domains require an input of sequential patterns, and some need both sequential input and output patterns. Among other approaches, Recurrent Neural Networks allow for persistent data across an arbitrarily long context-window unlike traditional feed-forward neural networks\cite{lipton2015critical}. However, this architecture can only handle even input size, while in many problems (such as language translation) the input lengths do not satisfy this requirement. Cho et al.\cite{cho2014learning} proposed a neural network with two RNNs, one being an \textit{encoder} and the other being a \textit{decoder}. This is referred to as a sequence-to-sequence method. The encoder encodes a sequence of symbols into a vector of fixed length, while the other RNN decodes the vector into another sequence of symbols. The two RRNs are trained to maximize the conditional probability of a target sequence when given a source sequence. Later, Sutskever et al.\cite{sutskever2014sequence} used two LSTM architectures for the encoder and decoder to increase the number of model parameters at negligible computational cost. Recently, Luong et al.\cite{luong2015effective} improved the model with an attention mechanism to selectively focus on parts of interest and achieved significant improvements on the translation problems. The weather based need prediction is similar to the translation problem which translates temporal weather sequential status into dynamic needs items during disasters. Therefore, we choose the sequence to sequence based model in our solution.

\section{The Need Forecasting Framework}
\begin{figure}[th]
 \centering
 \includegraphics[scale=0.42]{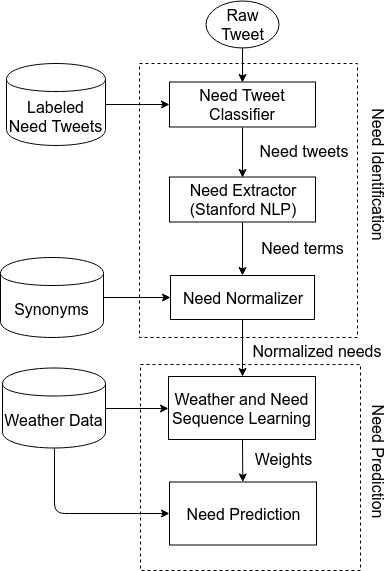}
 \caption{\label{fig:framework} The need prediction framework consists of two steps: 1: Need identification and extraction; 2: Need prediction.}
\end{figure}

The proposed forecasting framework is illustrated in Figure \ref{fig:framework}. Raw tweets will go through a need identification \& extraction step where need items are identified and extracted. Here we employ a supervised learning classifier to classify need tweets, based on which to extract need items using Stanford NLP~\cite{klein2003lex}, and then normalize need items. 
Once the current need items are available, they will be combined with weather data to feed into a sequence learning process to predict future need items. 

\label{sec:framework}

\begin{figure*}[t]
 \centering
 \includegraphics[width=\linewidth]{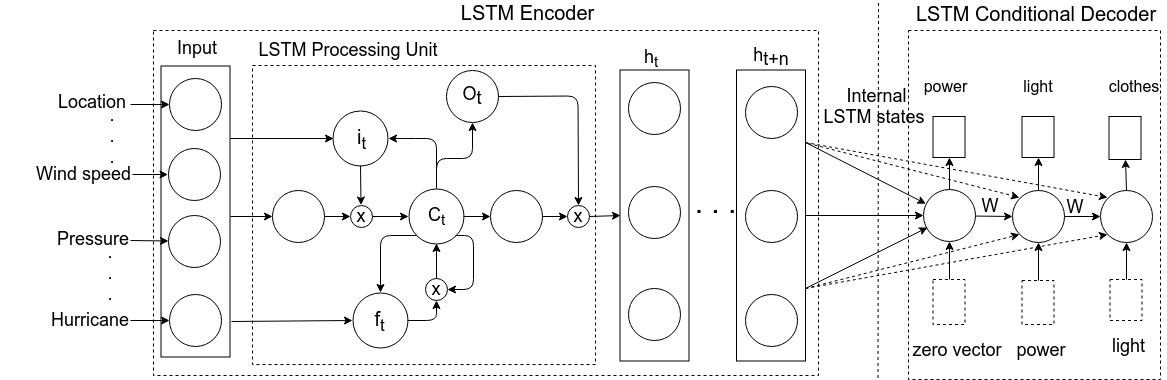}
 \caption{\label{fig:seq2seq} The need prediction sequence to sequence model. Encoder will encode the input to a fixed length vector representing the hidden state. Decoder will decode that hidden state to needs.}
\end{figure*}

\subsection{Need Identification and Extraction}
\subsubsection{Need tweet classification} A tweet that expresses one or more concerns is called need tweet. In the context of hurricane event, such a tweet usually contains one of the need terms such as \textit{help, diaper} or \textit{shelter, etc.}. 5,000 tweets were manually labeled as either need or none-need related tweets. We take 70\% of the dataset for training and the rest for test. A support vector machine (SVM) classifier was used to train and classify need tweets. Classical performance metrics such as \textbf{\textit{accuracy, precision, recall}} and \textbf{\textit{f-measure}} were used to evaluate the model. Our experiment demonstrated that the SVM classifier performances are \textit{0.932, 0.934, 0.939} and \textit{0.937}, in terms of these measures. This trained classifier is used to classify entire tweets in our dataset.

\subsubsection{Need extraction}
This step will extract features from tweets. In this case, the need items will be extracted. Usually, a need is represented as a noun. However, it can also be represented as a verb or adjective either explicitly or implicitly. To determine the part-of-speech (POS), the tweet must be analyzed by a natural-language-parser. In this case, we used the Stanford natural language parser\cite{klein2003lex} to give us the POS. After the POS are determined, the program extracts verbs, nouns, and adjectives, and we consider them as people's need or concerns. For example, suppose collected tweets are \textit{I need water at etc.}, \textit{My mom needs help etc.}, or \textit{she got trapped}. The resulting needs would be \textit{water, help} and \textit{trapped} respectively. Note that \textit{trapped} is not a need in usual sense. However, it is a situation that the person needs help. Therefore, we still include the adjective for need analysis. 
All other words and stopwords such as articles and event specific words like \textit{hurricane, Harvey or Houston etc.} are not considered because they are not needs semantically. Top $40$ most frequent words are extracted hourly as the most important needs requiring to be normalized in later phase.

\subsubsection{Need normalization}
To avoid duplication, need items are parsed through a normalization process by lemmatizing and transforming to their unique synonyms. In particular, we use Wordnet \cite{miller1998wordnet} - a Natural Language Processing (NLP) library to extend the words extracted from previous step into a number of different words based on its synonyms and lemmatization. Hence, each word provided from previous step will be extended to a set of words. And these words will play as candidates for the final normalized need word. To achieve the final normalized word, these words will be used to compare with a predefined need corpus. The matched word between the two sets is the normalized need word for later processing. In case of no matched word is found, these words will be manually evaluated before deciding to drop as not a need item or adding to the need corpus as a new need item. For example, given a corpus of a number of unique needs, the words \textit{donate, donates} and \textit{donation} are grouped to a common need in the corpus which is \textit{donation}. After the normalization, each word now represents single need. Repeated words are counted for the frequency. The final need corpus composes of $75$ different need items.

\subsection{The Sequence to Sequence Model}
\subsubsection{Long short term memory networks (LSTM)}
Recurrent neural networks have major issues regarding the vanishing or exploding gradient problems, which makes the networks hard to train. LSTMs address the problem by introducing a set of gating functions (units) to control the flow of information \cite{hochreiter1997long}.
As shown in the left box of Figure \ref{fig:seq2seq}, a common LSTM unit is composed of a cell $c_t$,  an input gate $i_t$, a forget gate $f_t$, and an output gate $o_t$. These three gates regulate the flow of information into and out of the cell, and  the cell remembers values over arbitrary time intervals.
The input gate $i_t$ takes activation from input layer at current time step and from the hidden layer at previous time step $h_{t-1}$. It is called a gate because its value is used to multiply with the value that flows through it.
with the sense that it will block the flow from other node if the gate value is zero. On the other hand, if its value is one, all the flow of values is allowed to pass to the cell.
The forget gate $f_t$ was introduced by Gers et al. in 2000 \cite{gers1999learning} which is used to flush the contents of the internal state.
The output gate $o_t$ controls the extent to which the value in the cell is used to compute the unit output activation.
Below is the summary operation of each update:
\begin{equation}
i_t = \sigma(W_{xi} x_t + W_{hi} h_{t-1} + W_{ci} c_{t-1} + b_i) 
\end{equation}
\begin{equation}
f_t = \sigma(W_{xf} x_t + W_{hf} h_{t-1} + W_{cf} c_{t-1} + b_f) 
\end{equation}
\begin{equation}
 c_t = f_t c_{t-1} + i_t tanh(W_{xc} x_t + W_{hc} h_{t-1} + b_c) 
\end{equation}
\begin{equation}
o_t = \sigma(W_{xo} x_t + W_{ho} h_{t-1} + W_{co} c_{t} + b_o )
\end{equation}
\begin{equation}
h_t = o_t tanh(c_t) 
\end{equation}
Notice that $\sigma$ is the logistic sigmoid function. The weight matrix $W_{hi}$ represents a hidden input gate matrix, $W_{xo}$ represents the input-output gate matrix while $W_{ci}$ represents a cell to gate matrix consisting of diagonal vectors. Detail can be referred at \cite{srivastava2015unsupervised}\cite{graves2013generating}.

\begin{figure}[th]
 \centering
 \includegraphics[scale=0.43]{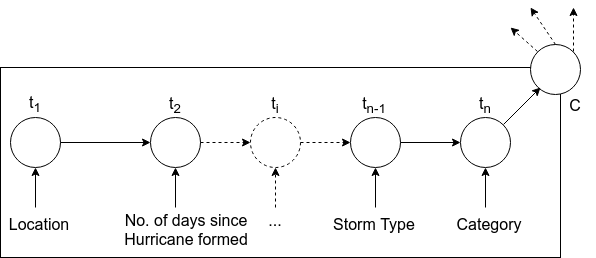}
 \caption{\label{fig:input-sequence} The input sequence designed to incorporate spatial temporal features. Each node represents one input symbol per timestep $t_i$. $C$ is the summary produced by the encoder.}
\end{figure}

\begin{figure*}[t]
	\centering
	\subfigure[Harvey]{
	    \includegraphics[scale=0.24]{figures/tbd/harvey.png}
	    \label{word-cloud-harvey}

	}
    \subfigure[Sandy]{
		\includegraphics[scale=0.18]{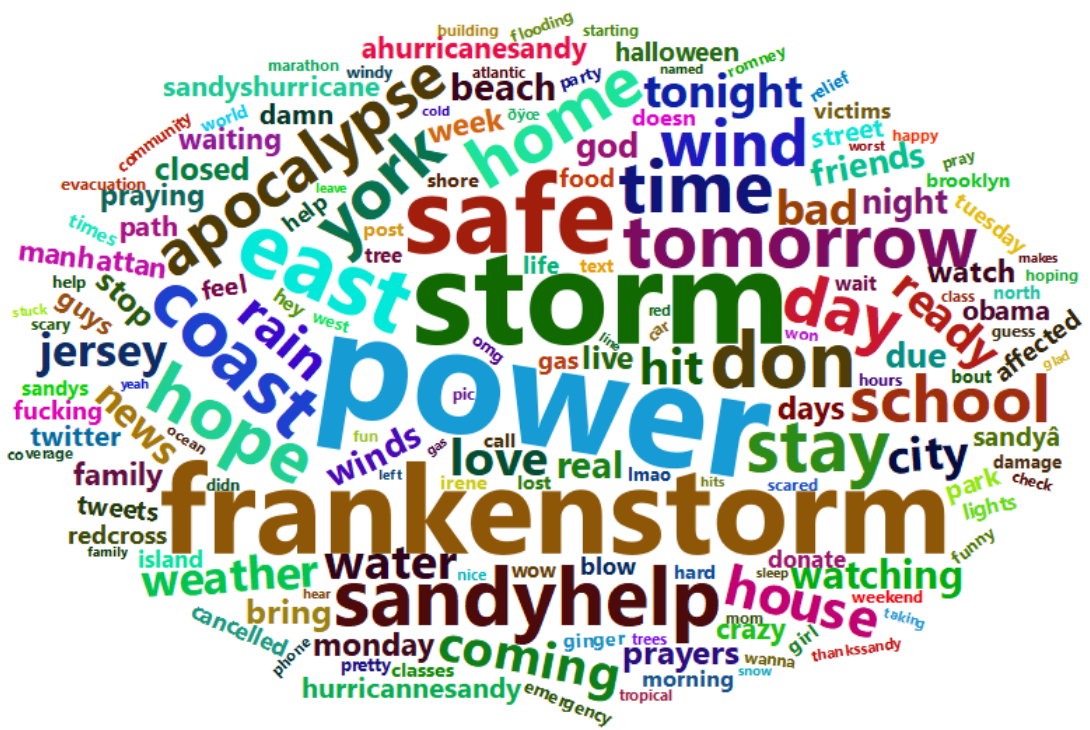}
		\label{word-cloud-sandy}
    }
    \subfigure[Irma]{
		\includegraphics[scale=0.15]{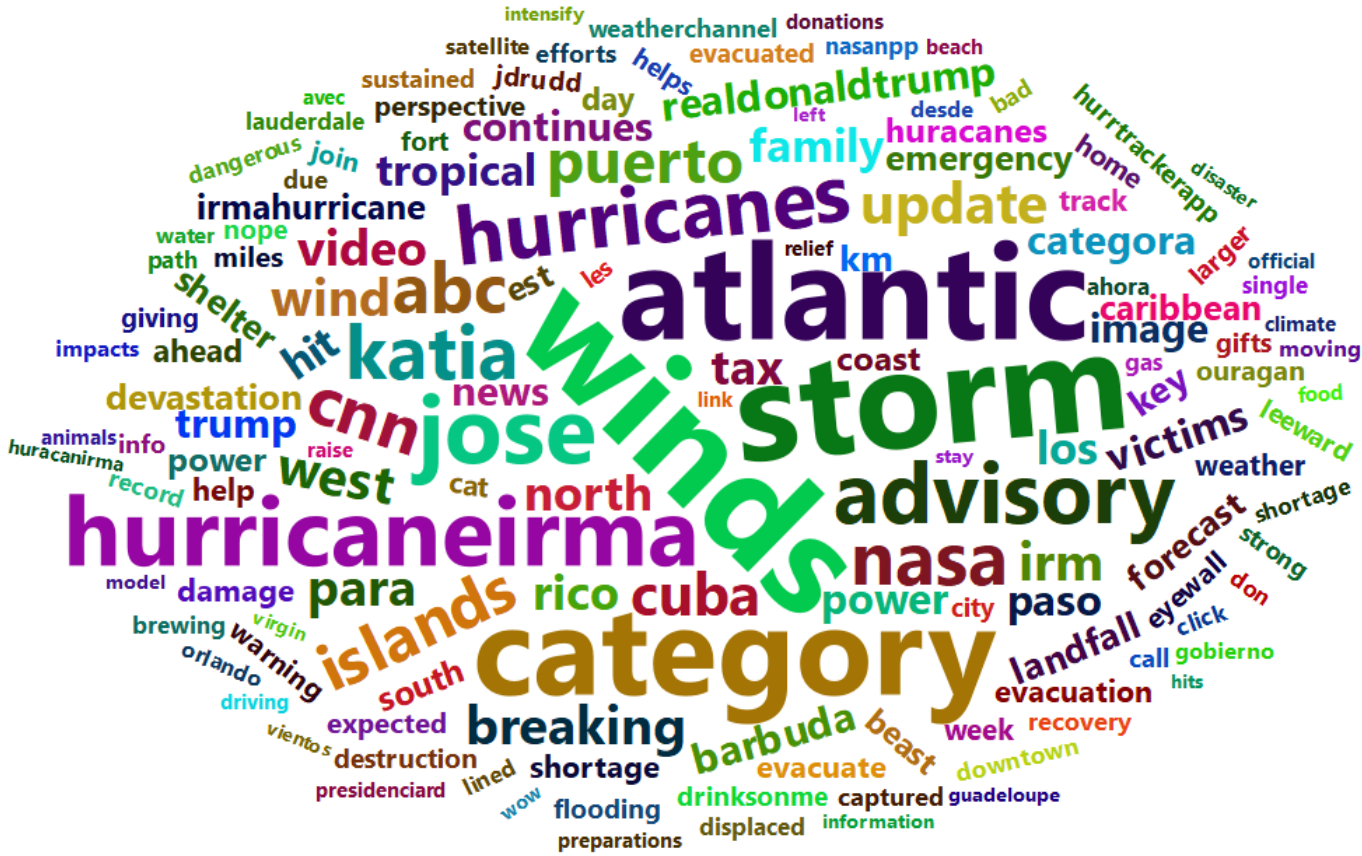}
		\label{word-cloud-irma}
    }
    \subfigure[Top Harvey Needs]{
	    \includegraphics[scale=0.46]{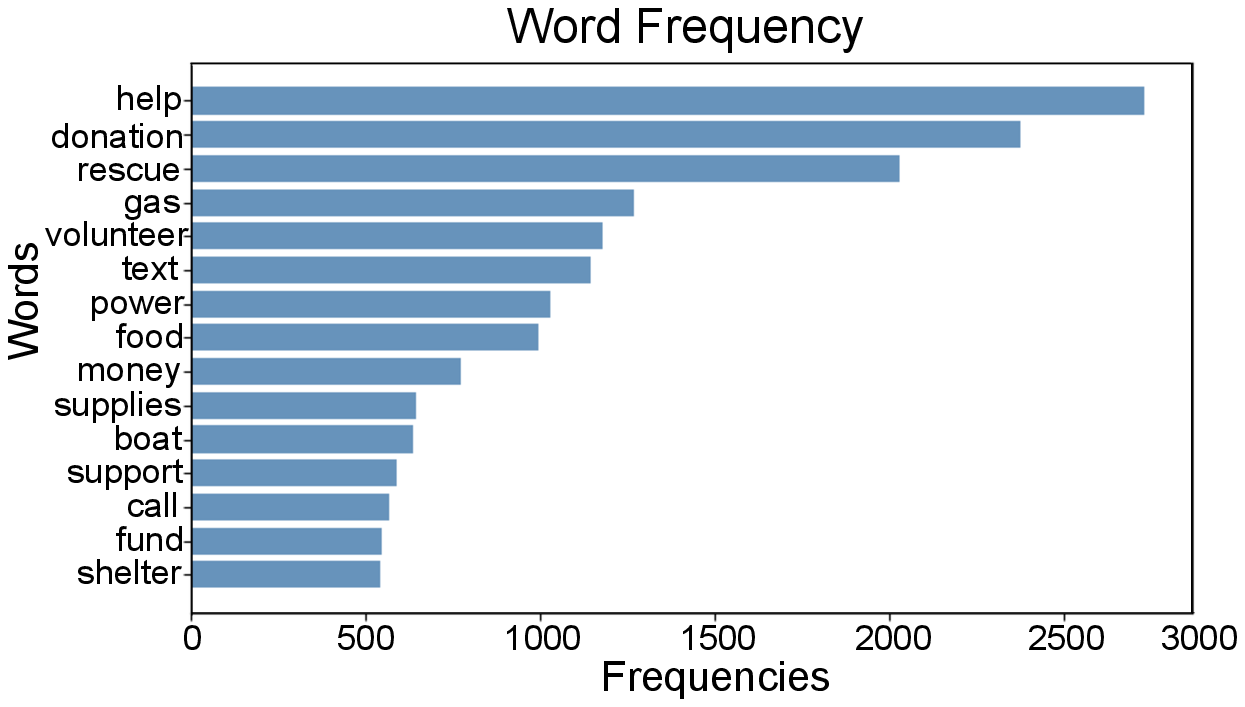}
	    \label{harvey-needs}

	}
    \subfigure[Top Sandy Needs]{
		\includegraphics[scale=0.48]{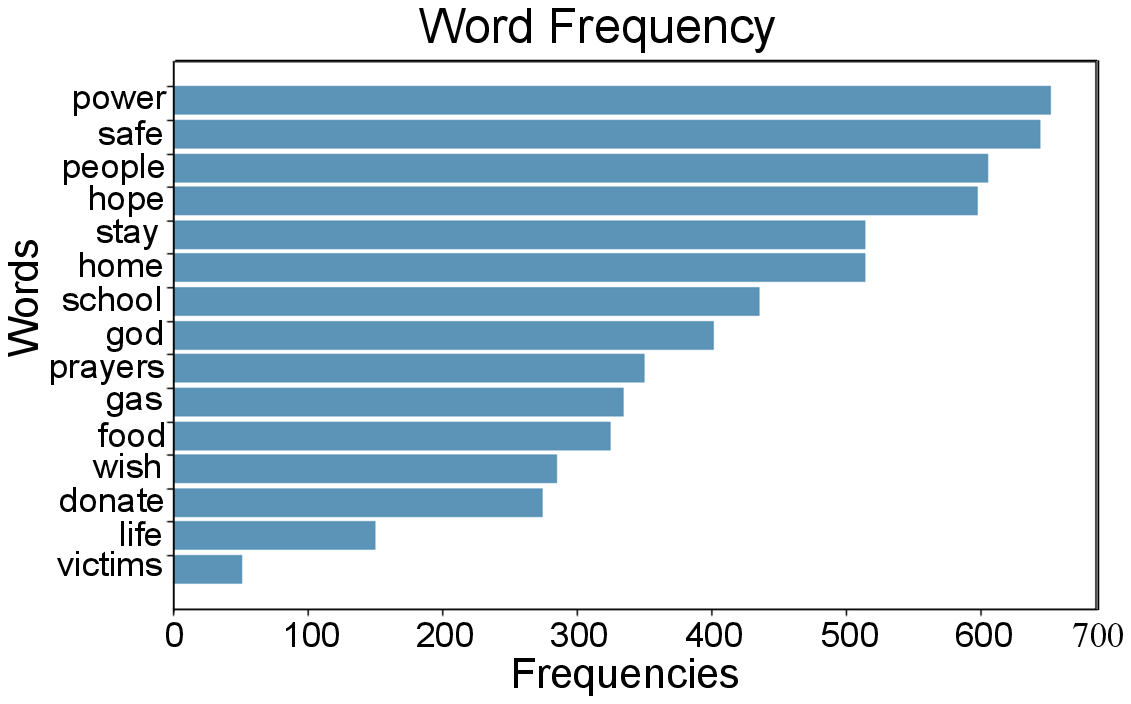}
		\label{sandy-needs}
    }
    \subfigure[Top Irma Needs]{
		\includegraphics[scale=0.48]{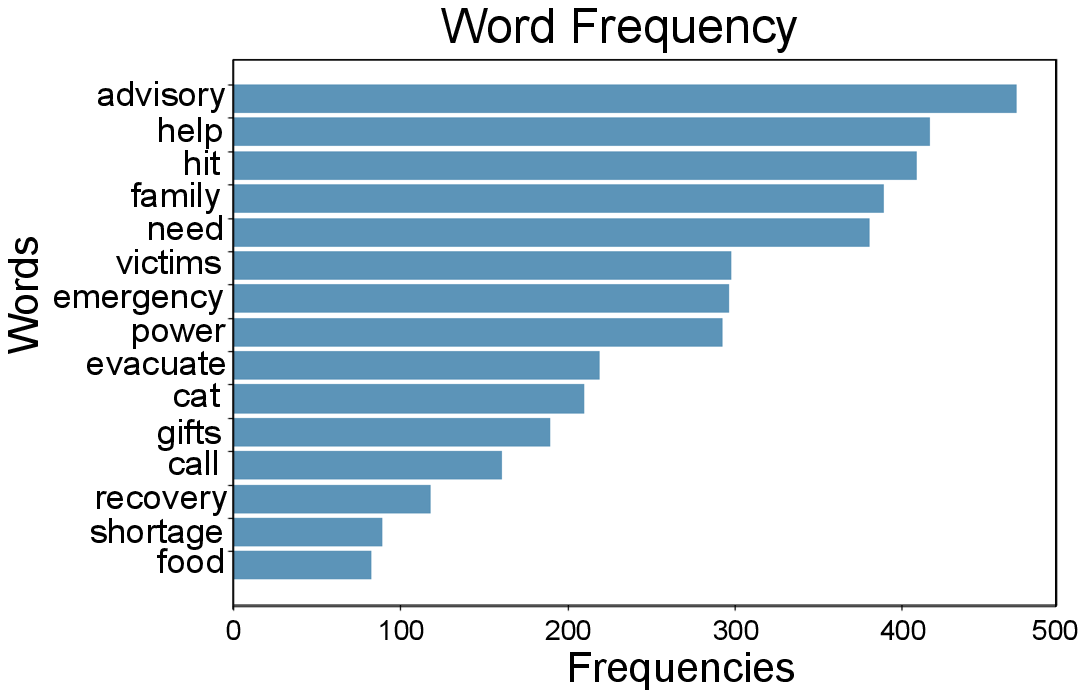}
		\label{irma-needs}
    }
	\caption{Word clouds and top needs in the three Hurricane events.}
\label{fig:hurricane-summary}
\end{figure*}

\subsubsection{Design of input and output to incorporate spatial-temporal features}
\label{subsub:design-input-output}

Figure \ref{fig:input-sequence} depicts our design of input sequence. The encoder reads each input symbol in one timestep, and it transfers the learned knowledge to the next timestep. After reading the end of of the sequence, the hidden state is a summary $C$ of the entire input sequence. This design helps to pass influences of prior factors to the subsequent ones. In our input sequence, location is chosen as the first symbol because it impacts the rest variables. The rest symbols are temporal features of the disaster such as the number of days since the Hurricane is formed, the current hour block. Lastly are weather features such as pressure, wind speed, storm type and the storm category. 

In addition, we treat all variables as categorical variables. Numerical variables such as $pressure$ and $wind speed$ are converted to categorical values. For example, $wind speed$ which has value range from $15(mph)$ to $200(mph)$, will take a range of $5(mph)$ as one category. The $pressure$ variable has value range from $950(mb)$ to $1020(mb)$. We use $1(mb)$ unit difference as one category. This makes each input symbol like a word in machine translation model. The total number of unique words in our vocabulary is $715$. This value is not too big; therefore, one-hot encoding can be utilized as the internal embedding mechanism.

The output sequence is the extracted needs of each hour block in temporal order regarding the tweet timestamps. If there are more than one need item in the same tweet, alphabetical order is chosen. This alignment guarantees the output sequence matched the conditional output in the decoder exactly.

\subsubsection{Need prediction}
Sequence to sequence learning is a way to convert data from one domain to another. Most often this is used for translating natural language sentences in a given language to a target language. When performing the prediction, the trainer will use the weather data as an input sequence and the need data as the target sequence. During training, the model tries to predict the target element based on the source element. As training continues, the weather data will become more coupled with the corresponding need. Figure \ref{fig:seq2seq} presents the concept of our sequence to sequence model. For demonstration, given the input sequence: location is ``Houston''; weather data for date ``Aug 26''; ``9'' days after Hurricane formed; the wind speed is ``115'' (mph); the pressure is 950 (mb); the storm type is ``hurricane'' and the category of this weather is ``3'', a numerical vector is created and passed through the encoder. Internal LSTM states of the encoder are achieved with a fixed dimension and then passed to the decoder together with its previous prediction. In each time step, the decoder will generate a need word until the special token ``[EOS]'' is found. This results in a final need output sequence ``power light clothes [EOS]''. The special token signifies the end of the predicted sequence.

\begin{figure}[t]
	\centering
	\subfigure{
	    \includegraphics[scale=0.29]{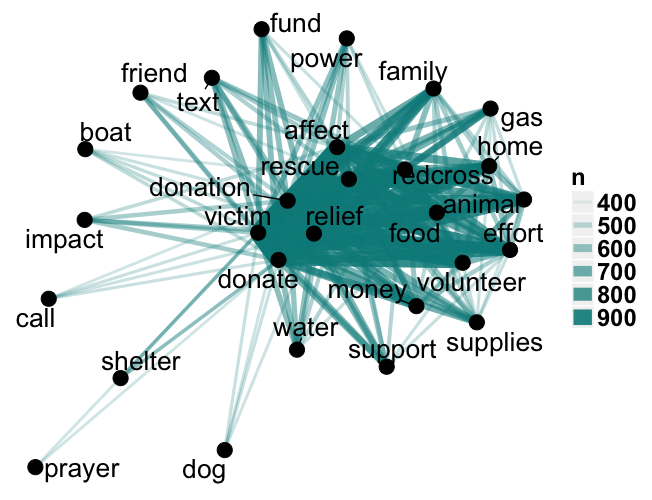}
	}
    \subfigure{
		\includegraphics[scale=0.30]{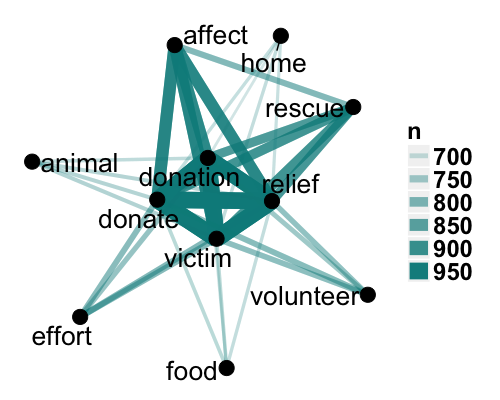}
		\label{wordnetwork2}
    }
	\caption{Word co-occurrence network with different frequency levels for Hurricane Harvey related needs (middle: $n\geq 400$; right: $n\geq 700$, where $n$ stands for the total number of concurrency of a word pair).}
\label{fig:tweet-coocurrence}
\end{figure}

In the model, there are two LSTMs involved. One serves as an encoder and the second serves as a decoder. 

\begin{itemize}[noitemsep,nolistsep]
\item \textit{The encoder}: This is an LSTM architecture that encodes an input sequence to its internal state with equal length and discard the output. This internal state plays as the context of the decoder in the second step.

\item \textit{The decoder}: This is also an LSTM architecture that decodes the context received from encoder, and its output from previous timestep to output sequence for current timestep.
\end{itemize}

The encoder takes a sequence of spatial, temporal and atmospheric symbols as input and produces a fixed size hidden states. Its internal states are then returned as inputs for the decoder. Each LSTM has $256$ neurons in the hidden layer and was trained in $100$ epochs. Beside the encoder's internal state, the decoder also takes the prediction at one time step earlier as inputs for its current prediction. This reflects the actual behavior of hurricanes where its weather condition changes quickly across time blocks. The goal of the LSTM is to estimate the conditional probability $p(y_{1},...y_{T^{'}}|x_{1},...x_{T})$ where $(x_{1},...x_{T})$ is an input sequence, $(y_{1},...y_{T^{'}})$ is its corresponding output sequence whose length $T^{'}$ may differ from $T$, and $v_t$ is the fixed dimensional representation of input sequence $(x_{1},...x_{T})$ given by the last hidden state of the LSTM. The conditional probability can be computed as \cite{NIPS2014_5346}:
\begin{equation}
\label{eq:lstm-prob}
p(y_{1},...y_{T^{'}}|x_{1},...x_{T})=\prod_{t=1}^{T'}p(y_{t}|v_t,y_{1}...y_{t-1})
\end{equation}
To make the decoder weigh differently on each part of the encoder's output, an attention mechanism is also incorporated in the decoder network. The attention mechanism combines input symbols and their weights in the decoder to perform the prediction. In other words, the context vector $v_t$ is computed in such a way that it can weigh specific parts of the input sequence. 
In particular, the attention weights are first computed using concatenation $ e_t = i_t \oplus  \overline{h}_{t} $ of the decoder's input $i_t$ and the hidden state $\overline{h}_t$ as in equation (\ref{eq:attn-weight}). Then it is multiplied with encoder's output vector to create weighted combination in equation (\ref{eq:attn-applied}).
\begin{equation}
\label{eq:attn-weight}
\alpha_{tj} = \frac{exp(e_{tj})}{\sum_{k}{exp(e_{tk})}}
\end{equation}
\begin{equation}
\label{eq:attn-applied}
v_t = \sum_{k}{\alpha_{tk}o_k}
\end{equation}
Therefore, the result shall contain information about the specific part of the input sequence, thus enhance the performance of the decoder to output the right needs. 

Overall, the encoder and the decoder are jointly trained to maximize the conditional probability of a target sequence given source sequence\cite{cho2014learning}.

\section{Experiments and Results}\label{sec:experiment}
\begin{figure}[t]
    \centering
    \includegraphics[scale=0.44]{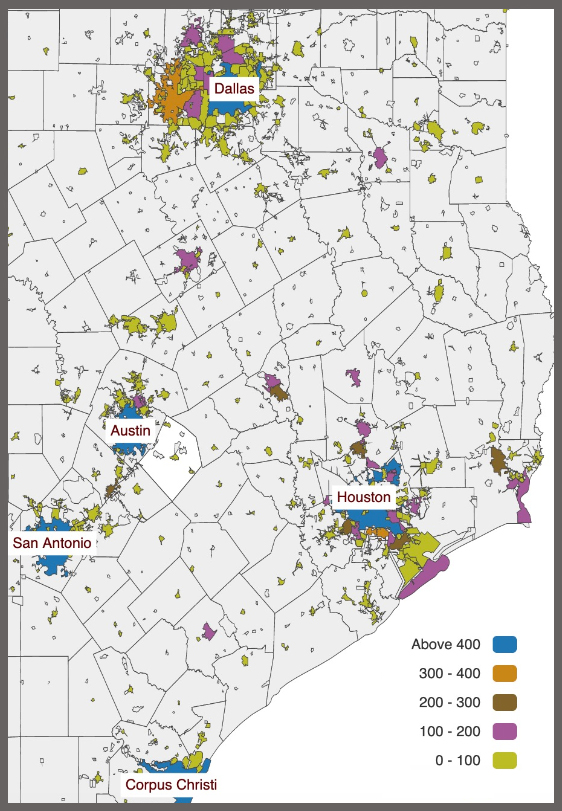}
    \caption{Need heat map by cities. Five big cities are presented as examples. The colors represent quantities of needs by terms extracted from the tweets.}
    \label{fig:need-heat-map}
\end{figure}
\begin{figure}[t]
    \centering
	\includegraphics[height=5.6cm, width=8.5cm]{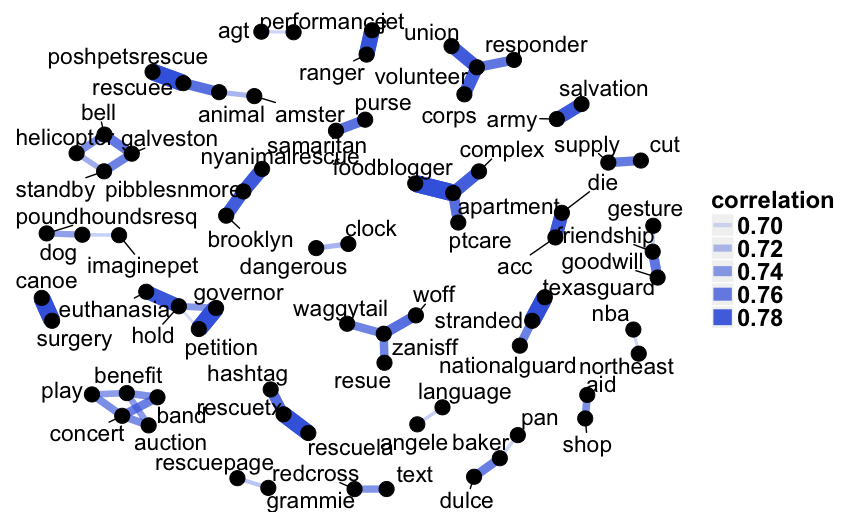}
    \caption{Word correlation for Harvey-related needs. The data is a slice of the whole dataset with the correlation coefficient between 0.7 and 0.8. The correlation indicates how often a word pair appears together relative to how often they appear separately. Phi coefficient is used here for binary correlation between a word pair.}
    \label{fig:word-correlation}
\end{figure}

\begin{figure}[t]
    \centering
    \includegraphics[height=4.6cm, width=8.5cm]{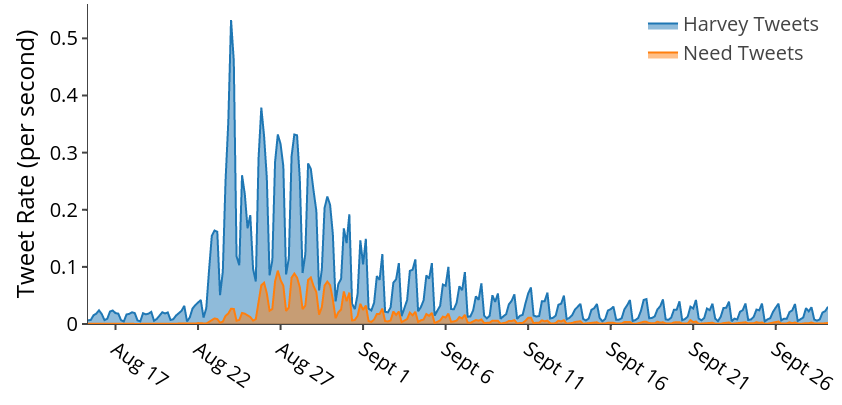}
    \caption{Hurricane Harvey related tweet rate and need-related tweet rate. The x-axis signifies the date, and each day is equally split into six time blocks.}
    \label{fig:tweet-rate}
\end{figure}

\subsection{Datasets}
The first dataset is Hurricane Harvey related tweets, which was collected from August 23 (right before hurricane arrival) to September 02, 2017 (right after the hurricane went off), with a total number of 140,000 tweets.
Hurricane Harvey related tweets are filtered out by keywords and hashtags such as `Harvey', `\#HarveySOS' and `HarveyRescue'. 
The weather dataset (wind speed, pressure and hurricane categories) was collected online, which is published by the Weather Company~\cite{weather}.

\begin{figure*}[th]
  \centering
  \includegraphics[width=17cm, height=6.4cm]{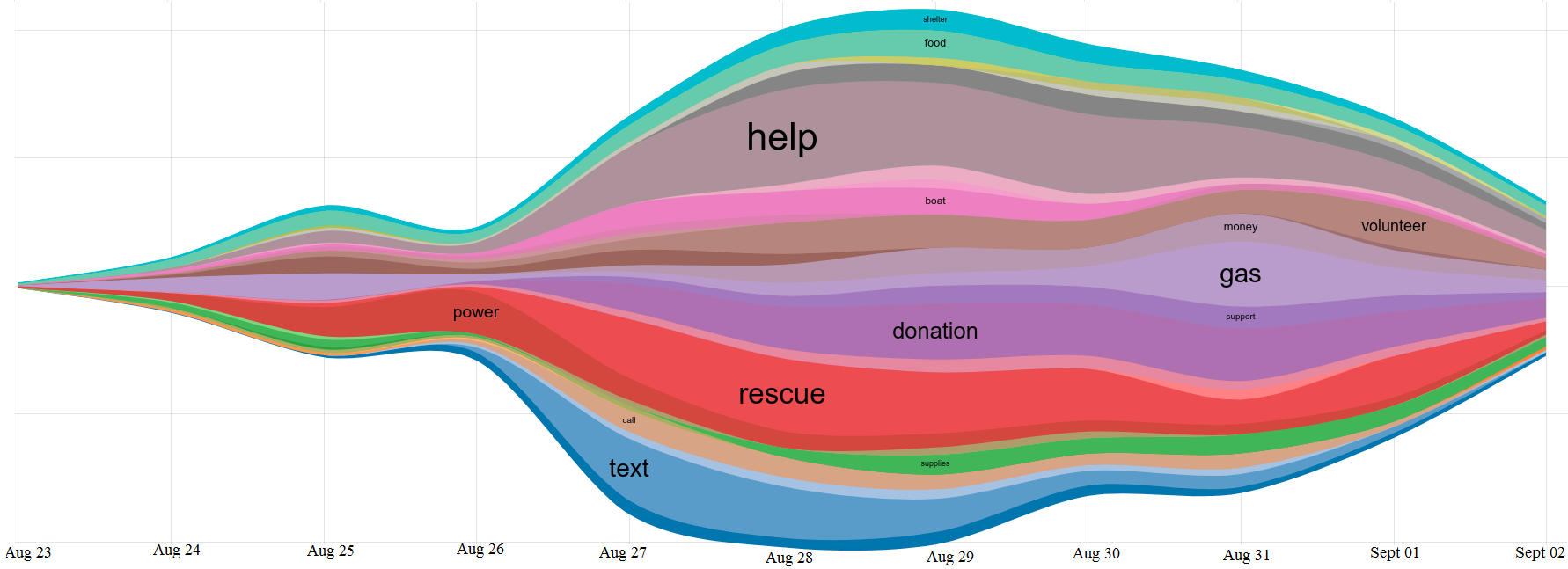}
  \caption{\label{fig:daily-need} Concern flow of the Hurricane Harvey from Twitter. The thickness of the area represents how much demanding of the concern is. The plot shows top $24$ concerns to make the text legible. Pick demanding period is from August 26 to August 29, 2017 when the Hurricane is at its deadliest impact. }
\end{figure*}

This study also uses tweets about the Hurricane Irma and the Hurricane Sandy to validate our prediction model. The Hurricane Irma dataset contains $60,000$ tweets from September 04, 2017 to September 13, 2017. The Hurricane Irma made landfall on September 09 in Florida and caused about \$$64.76$ billion loss. The Hurricane Sandy dataset contains $150,000$ tweets from October 24, 2012 to November 02, 2012, whose most destructive time was on October 29, 2012 when it reached the largest wind diameter at about $1,000$ miles. We use hourly data for training and prediction. Once features are extracted and prepared as described in section \ref{subsub:design-input-output}, we do random sampling and extract $80\%$ of the dataset for training and the rest $20\%$ for testing purpose.

\subsection{Needs Overview}
Here we show some interesting discoveries of the three hurricane event needs. Figure \ref{fig:hurricane-summary} summarizes the word clouds and top needs mined from tweets during the three hurricanes. Necessities such that \textit{``power'' and ``food''} existed in  all the three Figures of \ref{harvey-needs}, \ref{sandy-needs}, \ref{irma-needs}.
In Figure \ref{harvey-needs}, we can see that during Hurricane Harvey, \textit{``help'', ``donation''} and \textit{``rescue''} were the most frequent ones, which represented people's urgent and important needs. In addition, the high volume of \textit{``volunteer''} means that volunteers played a critical role when disaster Harvey happened. In Figure \ref{irma-needs}, we can see that during Irma, \textit{``advisory''} has the highest volume. This is probably because when Irma happened, people were more prepared than before, and the National Hurricane Center updated its status frequently with messages like \textit{``Hurricane \#Irma Advisory No. X''}. Hence, these tweets combined with people's retweets made \textit{``advisory''} the most frequent word. Similarly, there were a lot of mandatory evacuation orders from the government when Irma happened, making \textit{``evacuate''} another frequent word, which was different from Hurricane Harvey. In Figure \ref{sandy-needs}, we see some other frequent words such as \textit{``hope'', ``god''} or \textit{``wish''}. This is probably because Hurricane Sandy was the deadliest and most destructive hurricane in 2012, and had a strong psychological influences on people. Figure \ref{fig:need-heat-map} presents the number of needs by cities posted by people during Hurricane Harvey. We can see that Houston and Dallas are the two big cities that people tweet more about their concerns than other surrounding areas.

Word co-occurrence networks are used to examine which words commonly occur together. Here we take Hurricane Harvey as a case study. Figure \ref{fig:tweet-coocurrence} shows some strong connections in this network of words; words in the same time blocks with the high co-occurrence rates tend to go together. Connections can be visualized at different frequency levels. Despite the strong connection between words such as ``victim'', ``rescue'', ``relief'', ``donation'' and ``rescue'', clear clustering structure in the network is not observed. 
In Figure \ref{fig:word-correlation}, we also present the word correlation for Harvey-related needs whose correlations are between 0.7 and 0.8. This figure shows that other than general disaster-related terms such as ``responder'', ``volunteer'' and ``union'' which are clustered together, several more specific terms also emerge, e.g. ``waggytail'', which is an animal rescue organization (https://www.waggytailrescue.org/). Such detailed information will allow us to better understand the context of certain needs and corresponding population.

Figure \ref{fig:tweet-rate} shows an example of tweet rate in the event of Hurricane Harvey. It can be observed that when Hurricane Harvey formed on Aug 17, 2017, 
neither the Harvey related tweet rate or the needs related tweet rate has an obvious trend. As Hurricane Harvey marched toward Texas on Aug 23, there was a burst of Harvey related tweets, and also a sharp increase in needs related tweets. As time went by, the needs related tweet rate peak on Aug 27. As the influences of Hurricane Harvey decreased, the tweet rates in both categories decreased as well.

\subsection{Concern Flow}
Our system picks the most frequent need words in each day during the hurricane period and plots the concern flowchart. Figure \ref{fig:daily-need} shows an example of people's concern during Hurricane Harvey. Each word represents a type of need or concern from the public. The graph represents repeated concerns over time. The more discussed concerns are presented relatively proportional with the thickness of represented area in the graph.

As can be seen in Figure~\ref{fig:daily-need}, the top three dominated needs are \textit {help, donation} and \textit{rescue}. Two days before the arrival of the Harvey, we can see that people started discussing \textit{food, power} and \textit{supplies}. These are items that people need to prepare before the hurricane arrives. Along with previously mentioned needs people also discussed \textit{gas} which would be affected by the coming storm. Upon the Hurricane arrival on Aug 25, people started mentioning the topics \textit{power, pet, food} and \textit{house}. Notably, the need \textit{rescue, help, trap} and \textit{donation} were increasing from Aug 27 which clearly indicates the impacts of the devastation. We also observed the need \textit{text, call} and \textit{volunteer} that includes ways for people to contact for support. From Sept 2, there were decreasing trends of needs \textit{rescue, donation} and \textit{volunteer}. Their frequencies tended towards the frequencies of \textit{supplies, gas} and \textit{fund}. This indicates the hurricane over and maintains a demanding support for recovery.
On the other hand, we can also see variations of need priorities over different days. \textit{Supplies, power} and \textit{gas} are more needed before hurricane arrival. \textit{Rescue} and \textit{help} were the most important needs when the hurricane arrived, and \textit{donation} and \textit{volunteer} were the most mentioned terms when the hurricane faded out.

\begin{table*}[htbp]
\small
\caption{Samples of Result comparison of need prediction in Houston on Harvey dataset in different days and different hour blocks. S1, S2, S3 are the SMC of the Seq. to seq., LSTM and N-gram models respectively. The hour is hour block from minute $00$ to minute $59$. The result of CNN based model is not included due to limited space.}
\centering
\begin{tabular}{| p{1cm} | p{3cm} | p{3cm} | p{0.4cm} | p{3cm} | p{0.4cm} | p{3cm} | p{0.4cm} | }
\hline
\textbf{Date Time} & \textbf{Actual Need} & \textbf{Seq. to Seq. Predicted Need} & \textbf{S1} &
\textbf{LSTM Predicted Need} & \textbf{S2} &
\textbf{N-gram Predicted Need} & \textbf{S3} \\
\hline
Aug 23 22:00 
& hope update money call
& \textbf{money} \textbf{update} \textbf{call} 
& 0.98
& \textbf{hope} support water house \textbf{call}
& 0.93
& help shelter victim house stay cat \textbf{update} hit relief thing 
& 0.84
\\
Aug 24 16:00 
& stay school hit water house friend call update gas power
& \textbf{stay} \textbf{school} \textbf{hit} \textbf{friend call}
& 0.93
& \textbf{stay hit update power}
& 0.92
& help water god effort prayer affect shelter relief \textbf{gas power} care
& 0.80
\\
Aug 25 00:00
& gas care water god dog calm update stay
& \textbf{gas care god calm stay}
& 0.96
& \textbf{gas} house responder evacuate
& 0.87
& \textbf{gas care} rescue text emergency shelter house power \textbf{god} hope family shelter school support
& 0.81
\\
Aug 25 14:00 & stay water hit gas food house evacuate preparation &  \textbf{stay water hit gas house evacuate preparation} 
& 0.97 
& \textbf{stay hit evacuate house}  & 0.93 
& cat \textbf{stay water} thing effort shelter affect friend survive \textbf{house} god update & 0.81  
\\
Aug 26 09:00 & dog power god update prayer hope house life stay survive call shelter &  \textbf{dog god power update house life call survive} & 0.95 
& \textbf{dog} care \textbf{prayer} sleep \textbf{house}  & 0.83 
& stay water \textbf{house} call friend \textbf{hope} help \textbf{power} family prayer affect die gas care rescue text money & 0.76  
\\
Aug 26 22:00 & stay water house cat power die shelter family &  \textbf{stay water house cat} prayer \textbf{die shelter family } & 0.96 
& \textbf{stay house} text ship  & 0.88 
& help family call volunteer god survive water \textbf{stay house} relief update \textbf{water} victim affect donation thing friend support & 0.79  
\\
... & ... &  ... & ... & ...  & ... & ... & ...  
\\
... & ... &  ... & ... & ...  & ... & ... & ...  
\\
... & ... &  ... & ... & ...  & ... & ... & ...  
\\
Aug 29 17:00 & help rescue donation water life thing house relief boat shelter god power friend family food money animal die victim emergency effort volunteer &  \textbf{help relief life} donation stay \textbf{money} aid shelter rescue \textbf{house} call thing & 0.79 & help affect \textbf{food victim}  & 0.73 
& rescue hit \textbf{house} die family shelter support dog \textbf{food} care aid effort hit  \textbf{money emergency} call \textbf{donation} text \textbf{volunteer}  affect & 0.75  

\\

Aug 30 11:00
& help relief update house affect rescue life friend hit evacuate shelter boat victim donation stay volunteer support
& \textbf{help relief life stay rescue} prayer family \textbf{evacuate shelter boat support }
& 0.87
& \textbf{help} victim \textbf{friend} house \textbf{affect} call recovery
& 0.84
& \textbf{help} victim \textbf{affect shelter} stay family house prayer money hit \textbf{friend} god \textbf{evacuate} cat school emergency
& 0.79

\\
Aug 30 13:00
& help update affect house water family donation thing friend stay volunteer hit survive power school
& \textbf{help update house water thing stay volunteer hit friend} 
& 0.92
& \textbf{help} victim \textbf{affect house water family donation thing}
& 0.87
& \textbf{help} victim \textbf{donation update} recovery \textbf{family} relief hope call prayer support text thing
& 0.73

\\
Aug 31 21:00
& help recovery family relief volunteer hit victim update god thing school effort water house life
& \textbf{help relief god volunteer victim family thing effort life}
& 0.92
& support \textbf{help} boat rescue
& 0.77
& \textbf{water} stay cat \textbf{family} evacuate hope friend \textbf{god} prayer \textbf{thing life house}
& 0.80
\\
Sept 02 06:00
& house call relief prayer evacuate help volunteer support
& \textbf{house call relief prayer evacuate help} affect
& 0.96
& \textbf{house relief prayer} thing effort fund recovery
& 0.88
& \textbf{house relief} god medical
& 0.88
\\

\hline
\end{tabular}
\label{table:need-prediction}
\end{table*}

\subsection{Need Prediction}
\subsubsection{Baseline models}
\paragraph{\textbf{N-gram}}
An n-gram is a n-tuple or group of n words or characters (grams, for pieces of grammar) which follow one another \cite{jurafsky2014speech}. With this structure, the n-gram model captures the language structure from statistical point of view. It tells what letter or word is likely to follow a given one. In our study, weather information and the needs are combined to form word sequences. Each word is separated by a blank space. Each sequence is separated by a full stop. These sequences are used to build the n-gram model. From the given weather information, the n-gram model will keep predicting the next word and use it as the input for next prediction. This process is repeated until a symbol of sequence end is found (the full stop). Output words are predicted needs. To avoid be overfitted with the context if n is too big or not being able to capture general knowledge when n is too small, we choose n equals 3 for our base model.

\paragraph{\textbf{Generative LSTM}}
In this base model, the use of LSTM is changed to text generation. It is no longer a small unit as in the sequence to sequence architecture that it produces hidden states for the decoder. Instead, the LSTM layer connects directly to a dropout of 0.2 (to avoid overfitting and memorization) and a dense layer to produce its prediction. Onto the next word prediction, the same idea of the n-gram model is applied. The model will keep generating next words until the end of sequence is found in the output. 

\paragraph{\textbf{Convolutional Neural Network (CNN)}}
CNN based model has been popular recently in the application of text classification. We utilize the model proposed at \cite{kim2014convolutional} and simplified by \cite{bhaveshoswal} in our problem with multi-label classification setting. It comprises of an embedding layer, three convolution layers, three max pool layers and a dropout layer before feeding to the final dense layer with softmax function activation. The foretasted needs will be the top $K$ predicted classes based on probabilities, where $K$ is the maximum number of needs per training sample. 

\subsubsection{Evaluation metric}
Given a set of predicted needs \textbf{A}, and actual needs \textbf{B}, we convert these sets into binary representations. In this representation, the index that has value of one is the index of that predicted need in the need corpus. Other indices will have value of 0. We evaluate the prediction performance by calculating the simple matching coefficient (SMC). This coefficient will reflect both mutual absence and presence of needs in the corpus space. The SMC is computed by equation (\ref{eq-jaccard}):

\begin{equation}
\label{eq-jaccard}
\begin{split}
SMC(A,B) & = \frac{\mbox{Number of matching attributes }}{\mbox{Number of attributes}} \\
& = \frac{M_{00} + M_{11}}{M_{00} + M_{01} + M_{10} + M_{11}}
\end{split}
\end{equation}
where $M_{00}$ and $M_{11}$ are the total number of attributes where \textbf{A} and \textbf{B} both have $0$ and $1$, respectively. $M_{01}$ is the number of attributes where \textbf{A} is $0$ and \textbf{B} is 1; while $M_{10}$ is the number of attributes where \textbf{A} is $1$ and \textbf{B} is $0$. 
From now onward, the term \textit{similarity, score, accuracy,} and \textit{SMC} are used interchangeably. 

\subsubsection{Results and comparison}
Table \ref{table:need-prediction} shows a sample of predicted needs and its score using the three prediction methods for the hurricane Harvey event. In the table, \textbf{S1}, \textbf{S2} and \textbf{S3} represent computed score for Seq. to seq., LSTM and n-gram prediction model respectively. Each need is separated with a single space and the correct predicted needs are bold.

\begin{table}[t]
\setlength{\tabcolsep}{10pt}

  \centering
    \caption{Hourly average prediction scores on each day of various techniques on Hurricane Harvey dataset. Bold values represent better performance.}
    \begin{tabular}{l|c|c|c|c}
    \toprule
    Date\textbackslash Method    & \multicolumn{1}{l|}{Seq. to Seq.} & \multicolumn{1}{l|}{LSTM} & \multicolumn{1}{l|}{N-gram} & \multicolumn{1}{l}{CNN} \\
    \midrule
    Aug. 23    & \textbf{0.89} & 0.85  & 0.83  & 0.68 \\
    \midrule
    Aug. 24    & \textbf{0.89} & 0.86  & 0.82 & 0.69 \\
    \midrule
    Aug. 25   & \textbf{0.88} & 0.84  & 0.82 & 0.69\\
    \midrule
    Aug. 26   & \textbf{0.87} & 0.85  & 0.84  & 0.68 \\
    \midrule
    Aug. 27    & \textbf{0.89} & 0.83  & 0.79  & 0.69\\
    \midrule
    Aug. 28   & \textbf{0.87} & 0.80  & 0.78  & 0.70\\
    \midrule
    Aug. 29   & \textbf{0.86} & 0.80  & 0.79  & 0.70\\
    \midrule
    Aug. 30   & \textbf{0.86} & 0.81  & 0.78   & 0.69\\
    \midrule
    Aug. 31  & \textbf{0.88} & 0.86  & 0.79  & 0.69\\
    \midrule
    Sept. 01  & 0.86 & \textbf{0.87}  & 0.82  & 0.69\\
    \midrule
    Sept. 02  & \textbf{0.88} & 0.86  & 0.81  & 0.69\\
    \bottomrule
    \end{tabular}
    \label{table:daily-avg-score}
\end{table}

\begin{table}[t]
\setlength{\tabcolsep}{8pt}

  \centering
    \caption{Average need prediction accuracy by Hurricane.}
    \begin{tabular}{l|c|c|c|c}
    \toprule
    Hurricane\textbackslash Method    & \multicolumn{1}{l|}{Seq. to Seq.} & \multicolumn{1}{l|}{LSTM} &
    \multicolumn{1}{l|}{N-gram} &
    \multicolumn{1}{l}{CNN} \\
    \midrule
    Harvey    & \textbf{0.875} & 0.839  & 0.806 & 0.690 \\
    \midrule
    Irma    & \textbf{0.882} & 0.858  & 0.792 & 0.682\\
    \midrule
    Sandy   & \textbf{0.878} & 0.845  & 0.805 & 0.678\\

    \bottomrule
    \end{tabular}
    \label{table:hurricane-accuracy}
\end{table}

To compare the performance of each technique, we calculate its similarity score as specified in equation (\ref{eq-jaccard}). Similarity score is calculated hourly and everyday to measure average daily performance of each forecasting method. The result in Table \ref{table:daily-avg-score} indicates that the daily performance of the sequence to sequence is better than the LSTM model, and outperforms the n-gram, and CNN models on every day prediction. The CNN based model shows the worst performance. This might be because CNN uses top K probability to produce output, while other models use sequential generation strategy for prediction.

\begin{figure}
    \centering
    \includegraphics[width=\linewidth]{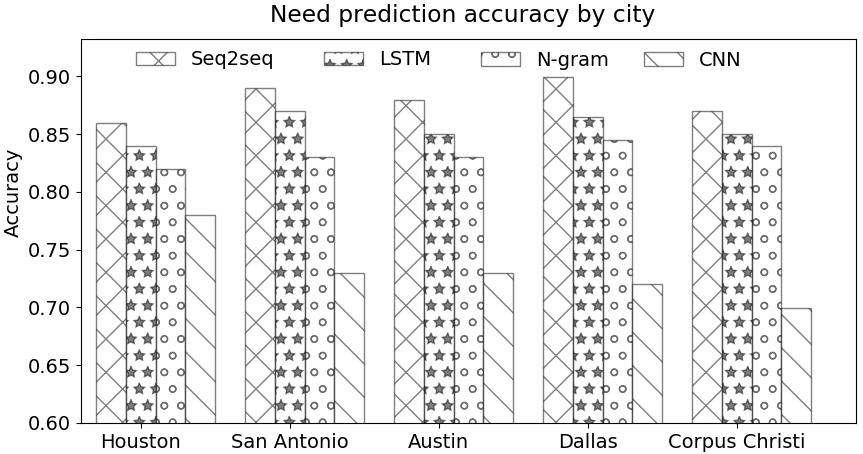}
    \caption{Need prediction accuracy of the proposed model by selected cities in Hurricane Harvey.}
    \label{fig:city-accuracy}
\end{figure}

In addition, we average need prediction by cities and show 5 cities from Hurricane Harvey dataset in Figure \ref{fig:city-accuracy}. The result shows that, the sequence to sequence model constantly outperforms LSTM, n-gram, and CNN models across cities.
We also average the prediction accuracy of all days' prediction in each hurricane. This will give an overall view of the accuracy of our proposed technique and other baselines across three datasets. This result is shown in Table \ref{table:hurricane-accuracy}. For the Hurricane Harvey, the proposed sequence to sequence model has the similarity score of $0.875$ while other three baseline models' scores are $0.839$, $0.806$, and $0.690$ for the LSTM, n-gram, and the CNN based model respectively. For the Hurricane Irma and Hurricane Sandy, the proposed Seq. to seq. model achieved similarity scores of $0.882$ and $0.878$ which clearly outperformed LSTM, N-gram, and CNN models. 
This result proves that our proposed technique is more reliable and works well with people's daily need prediction, which could help public agencies and other stakeholders be more prepared to natural disasters.
\subsubsection{Training loss and decoder's attention in prediction}
\begin{figure}
    \centering
    \includegraphics[scale=0.6]{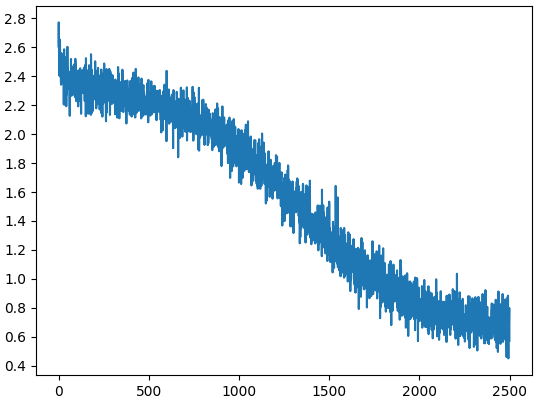}
    \caption{Training loss of the proposed model after $250,000$ iterations of training. X-axis represents number of iterations. Y-axis represents average loss per $100$ iterations.}
    \label{fig:training-loss}
\end{figure}

\begin{figure*}[th]
  \centering
  \includegraphics[width=\linewidth]{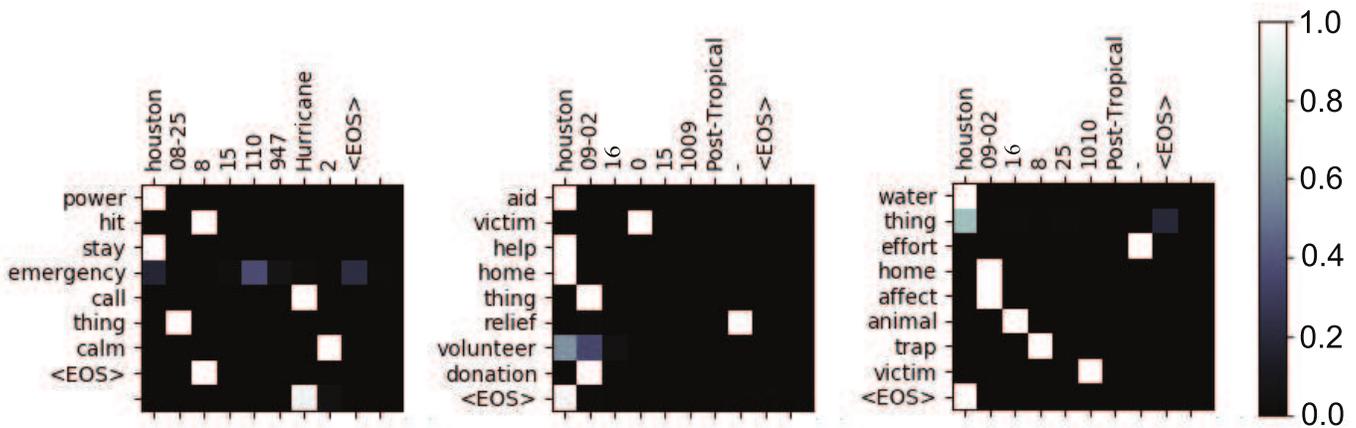}
  \caption{\label{fig:attention} Visualization of normalized attention weights in need prediction. The x-axis and y-axis of each plot correspond to the input features and decoded needs. The scale represents normalized weight that each input factor plays in the prediction.}

\end{figure*}

Figure \ref{fig:training-loss} presents the training loss of the sequence to sequence model in need prediction. The figure shows convergence trend of the model training loss from $2.8$ to $0.5$ of distance unit after $250,000$ iterations. Moreover, the loss decreases quickly before the iteration $200,000^{th}$ and gradually decreases afterwards. It shows that our model is well trained and has gained a certain accuracy level from the training set. In addition, it also shows the training is adequate due to the slow learning progress of the model after $200,000$ training iterations.

Once the model is trained, we perform need prediction and show three examples of the decoder's attention visualization in Figure \ref{fig:attention}. For example, with the input at location \textit{"houston"}, month and date block as \textit{"08-25"}, and \textit{"8"} days after the Hurricane formed, and other variables such as hour block, wind speed, pressure, storm type, storm category are
\textit{"15", "110", "947", "Hurricane", "2"} respectively; the model predicts \textit{"power", "hit", "stay", "emergency", "call", "thing", "calm"} as needs. From the figure, we see that the location always corresponding higher weight to needs such as \textit{"power", "aid"} and \textit{"water", etc.}. This can be explained as people in location $Houston$ faced power problem more often than other locations; they need more aid and bottle water due to the serious consequence of the Hurricane. The Hurricane type and Hurricane category have high impacts to needs such as \textit{"effort", "calm", "call", "relief"} since these variables represent the strength as well as the destruction of the Hurricane. 
\vspace{-2mm}
\subsubsection{The significance of the weather data in the proposed need prediction model}
\begin{figure}
    \centering
    \includegraphics[width=\linewidth]{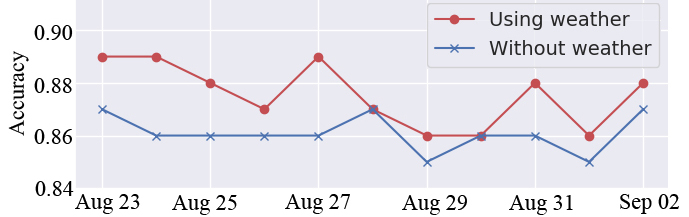}
    \caption{Hourly average need prediction accuracy comparison on each day during Hurricane Harvey.}
    \label{fig:hourly-average-each-day}
\end{figure}

In order to evaluate the significance of the weather information in the need prediction, we compare the performance of our proposed model with the model in which it does not include weather and only use previous needs as the input. The hourly average accuracy on each day is computed and shown in Figure \ref{fig:hourly-average-each-day}. We can tell that the model including weather data as input performs constantly superior than the model using only prior needs as input. The discrepancy between two models is decreasing from the beginning period of the Hurricane on August 23 until the final phase on September 02. It indicates that when the weather varies at the beginning and middle phases, it plays an important role in the need prediction. However, when the Hurricane is close to end in late August and early September, the weather information becomes less important and the two models reach to small discrepancy of prediction accuracy. 
\section{Discussion}\label{sec:conclusion}

In this paper, we present a novel sequence to sequence framework to forecast people's needs in disasters by exploring spatial, temporal and atmospheric factors altogether. This work is one of the first towards forecasting people's needs using massive social network data and weather data. To understand spatial temporal influence better, a careful alignment of the input sequence is designed. In addition, the sequence to sequence model is incorporated with an attention mechanism to identify the importance of each input variable in the prediction result. The case studies using data collected during Hurricane Harvey, Hurricane Irma and Hurricane Sandy show that people's concern flow can be tracked over time and top needs can be predicted more accurately than the LSTM generative, n-gram and CNN models. 
The proposed framework for need analysis and forecast is significant in that it provides an end-to-end interaction mechanism to respond to public's concern promptly, which is truly suitable for real disaster management.

\section{Acknowledgment}
This work was supported by the U.S. National Science Foundation under Grant CNS-1737634. 
\bibliographystyle{IEEEtran}
\bibliography{reference}

\vspace{-12mm}
\begin{IEEEbiography}[{\includegraphics[width=0.8in, clip,keepaspectratio]{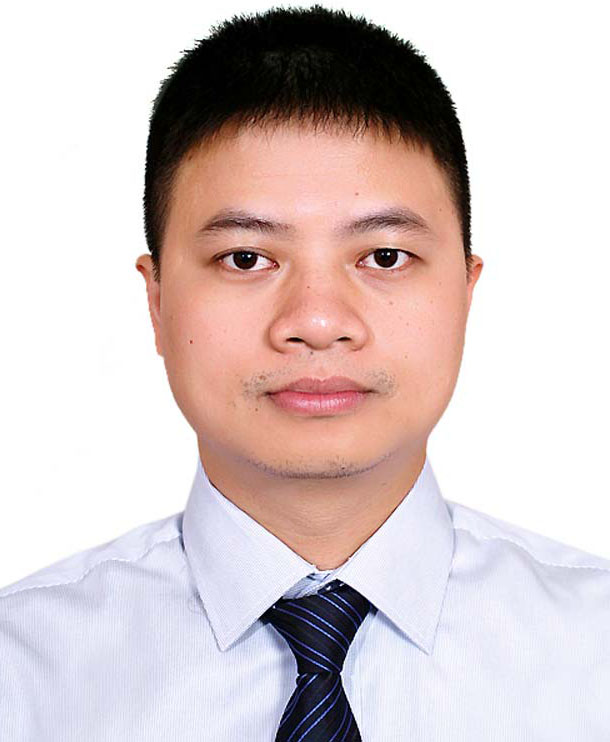}}]{Long Nguyen}
is a Ph.D. student in the Department of Computer Science at Texas Tech University. His research interests are in machine learning, data mining and data driven modeling for real world applications. He received his master degree from Politecnico di Milano (Italy) in 2009 and bacholor degree from President University (Indonesia) in 2005.
\end{IEEEbiography}
\vspace{-10mm}
\begin{IEEEbiography}[{\includegraphics[width=0.8in,clip,keepaspectratio]{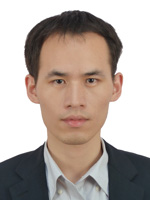}}]{Zhou Yang}
is a Ph.D. student in the Department of Computer Science at Texas Tech University. He received his bachelor’s degree at Wuhan University (China) in 2013, and he graduated from University of Florida with a master’s degree in Industrial Engineering in 2015. Before pursuing his Ph.D degree, he was a big data analyst in China Unicom. He has a great interest in machine learning and its application to industries.
\end{IEEEbiography}
\vspace{-12mm}
\begin{IEEEbiography}[{\includegraphics[width=0.8in,clip,keepaspectratio]{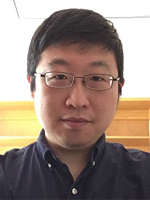}}]{Jia Li}
is a research assistant professor at TechMRT Center, Department of Civil, Environmental and Construction Engineering, Texas Tech University. He received his Ph.D. degree from University of California Davis in 2013, master degree from Hong Kong University of Science and Technology in 2008. He has great interest in traffic flow theory, queuing systems modeling and design.
\end{IEEEbiography}
\vspace{-11mm}
\begin{IEEEbiography}[{\includegraphics[width=0.8in,clip,keepaspectratio]{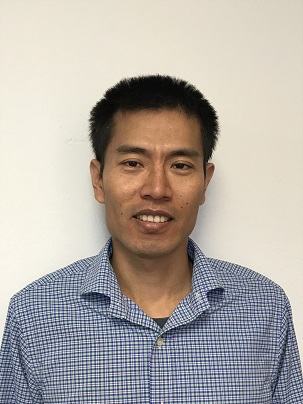}}]{Zhenhe Pan}
is a senior software engineer at kinetica db Inc in Arlington, Virginia. He received his first master degree of Computer Applied Technology in Chinese Academy of Science in 2007, and second master degree of computer engineering from Virginia Tech in 2014. He has passion in database design, optimization, and machine learning. 
\end{IEEEbiography}
\vfill
\vspace{-15mm}
\begin{IEEEbiography}[{\includegraphics[width=0.8in, clip,keepaspectratio]{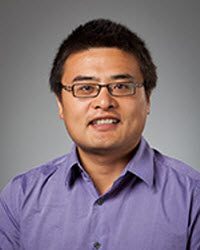}}]{Guofeng Cao}
is a GIS professor at Texas Tech University. His research primarily focuses on developing statistical methods and high performance computing methods to characterize and model spatial and spatiotemporal patterns. He was broadly trained in earth sciences, computer science, applied statistics, geographic information science and remote sensing.
\end{IEEEbiography}
\vfill
\vspace{-15mm}
\begin{IEEEbiography}[{\includegraphics[width=0.8in,clip,keepaspectratio]{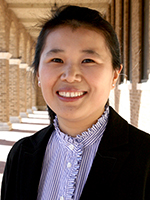}}]{Fang Jin}
is an assistant professor in the Department of Computer Science at Texas Tech University. She received her Ph.D. degree from The Department of Computer Science at Virginia Tech in 2016. She has a broad interest in Data Mining and Machine Learning. Her research has been focused on information propagation modeling, anomaly detection, and spatiotemporal data analysis.
\end{IEEEbiography}

\vfill

\end{document}